\pdfoutput=1

\documentclass[11pt]{article}

\usepackage[final]{coling}

\usepackage{times}
\usepackage{latexsym}

\usepackage{colortbl,array,xcolor}

\usepackage[T1]{fontenc}

\usepackage[utf8]{inputenc}

\usepackage{microtype}

\usepackage{inconsolata}

\usepackage{graphicx}
\usepackage{array}
\usepackage{tabularx}
\usepackage{cleveref}
\crefname{figure}{Figure}{Figures}
\usepackage{multirow,multicol}
\usepackage{hyperref}
\usepackage{xurl}

\title{Human-Like Embodied AI Interviewer: \\Employing Android ERICA in Real International Conference}

\author{
 \textbf{Zi Haur Pang},
 \textbf{Yahui Fu},
 \textbf{Divesh Lala},
 \textbf{Mikey Elmers},
 \textbf{Koji Inoue},
 \textbf{and}
 \textbf{Tatsuya Kawahara}
\\
  Graduate School of Informatics, Kyoto University, Japan 
  \\
  \texttt{\{pang, fu, lala, elmers, inoue, kawahara\}@sap.ist.i.kyoto-u-ac.jp}}

\begin{document}
\maketitle
\begin{abstract}

This paper introduces the human-like embodied AI interviewer which integrates android robots equipped with advanced conversational capabilities, including attentive listening, conversational repairs, and user fluency adaptation. 
Moreover, it can analyze and present results post-interview. We conducted a real-world case study at SIGDIAL 2024 with 42 participants, of whom 69\% reported positive experiences. 
This study demonstrated the system's effectiveness in conducting interviews just like a human and marked the first employment of such a system at an international conference. 
The demonstration video is available at \url{https://youtu.be/jCuw9g99KuE}.

\end{abstract}

\section{Introduction}

Qualitative interviews are foundational to social science research, offering deep insights through open-ended conversations. 
However, these interviews require considerable time and human effort. 
Earlier efforts to alleviate these demands involved using virtual agents \cite{nunamaker2011embodied, anderson2013tardis, sb2021improving}. 
Yet, these systems often failed to provide the sophisticated human-like interaction needed for quality research, limited to simple behaviors like head nodding and assuming participants' full understanding and fluent speech. 
This basic approach does not account for the complexities of real-world interactions, such as varied understanding and communication skills among participants, resulting in data quality and engagement shortfalls.

To address these limitations, this paper introduces a novel, \textbf{human-like interview system} that employs android and humanoid robots. 
This system is equipped with functionalities like advanced listening behaviors, conversational repair strategies, and user-fluency adaptation, which significantly enhance interaction quality. 
Beyond mere data gathering, our approach includes an end-to-end \textbf{post-interview processing workflow} where chained large language models (LLMs) handle data processing, analysis and presentation creation. 
We conducted a \textbf{real-world case study} at an international academic conference, where it facilitated numerous interactions, demonstrating its practical utility and efficiency. 
Notably, this marks the first instance of such a system being used at an international conference, showcasing our pioneering approach in the field. 
The comparative effectiveness of our system relative to traditional interview methodologies is detailed in Table \ref{tab:comparison}.

\begin{table*}[h]
\caption{Comparison of embodied AI interview systems. Bold highlights features unique to our proposed system.}
\label{tab:comparison}
\centering
{\small
\begin{tabularx}{\textwidth}{llXXc}
\hline
\multicolumn{1}{c}{System}                                         & \multicolumn{1}{c}{Agent} & \multicolumn{1}{c}{\begin{tabular}[c]{@{}c@{}}Agent Behavior\end{tabular}}  & \multicolumn{1}{c}{Dialogue Features}   & \multicolumn{1}{c}{\begin{tabular}[c]{@{}c@{}}Post-Interview \\Processing Workflow\end{tabular}} \\ \hline
\rowcolor[gray]{0.95} \begin{tabular}[t]{@{}l@{}}SPECIES\\ \cite{nunamaker2011embodied} \end{tabular}& Virtual & Eye Blink, Head Nodding & Follow-up Question & No \\ 
\begin{tabular}[t]{@{}l@{}}Maya\\ \cite{sb2021improving}\end{tabular}& Virtual & Gestures, Head Nodding & Follow-up Question & No \\ 
\rowcolor[gray]{0.95} \begin{tabular}[t]{@{}l@{}}ERICA\\\cite{inoue2021job}\end{tabular}& Robotic & Eye Blink, Lip Sync, Head Nodding & Follow-up Question & No \\ 
\begin{tabular}[t]{@{}l@{}}ERICA \& TELECO \\(Ours)\end{tabular} & Robotic & Eye Blink, Lip Sync, Gestures, Head Nodding, \textbf{Verbal Backchannel} & Follow-up Question, \textbf{Conversational Repair}, \textbf{User Fluency Adaptation} & \textbf{Yes} \\ \hline
\end{tabularx}
}
\end{table*}

\begin{figure*}[t]
  \centering
  \includegraphics[width=\linewidth]{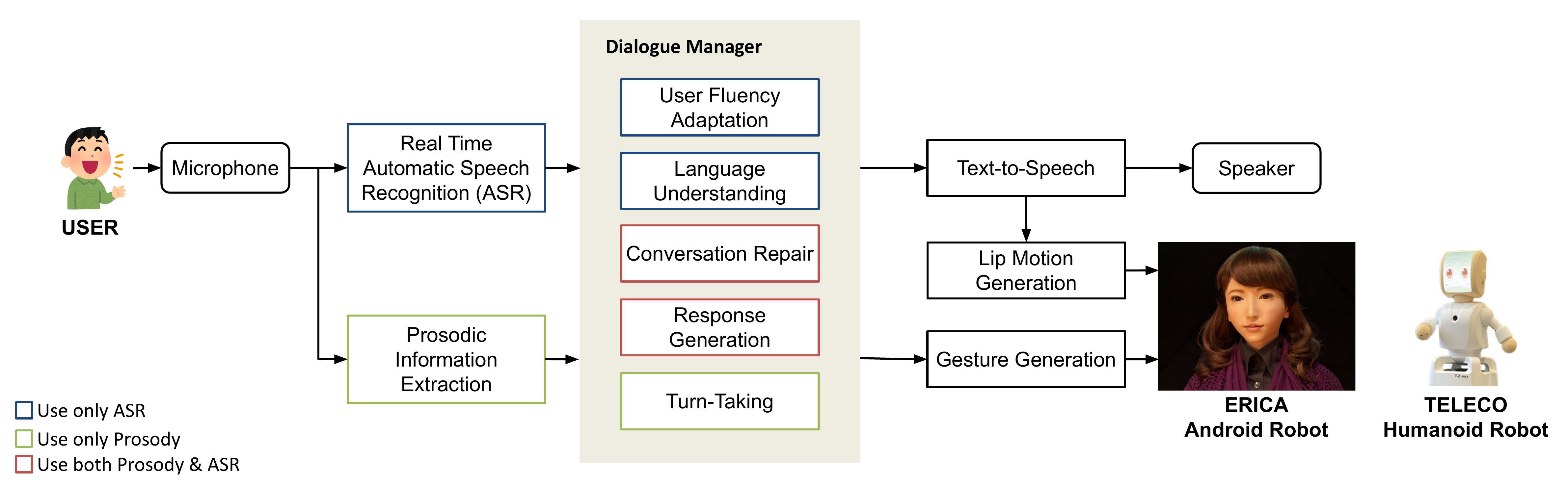} 
  \caption{Overall architecture of human-like interview system}
  \label{fig:architecture}
\end{figure*}

\section{Human-like Interview System}

In this section, we describe the architecture of our human-like interview system, as depicted in Figure \ref{fig:architecture}. 
The system initiates with a speech processing module that serves as the primary input mechanism. 
The core component, the dialogue manager, orchestrates tasks from language comprehension to response generation, including a Voice-Activity-Projection (VAP) based Multilingual Turn-Taking Module for effective turn management \cite{inoue2024multilingual, inoue2024real}. 
Additional features of the system, such as speech synthesis and gesture generation, are outlined in subsequent subsections. 
Following the discussion of these components, the interview dialogue flow and the post-interview processing workflow are detailed. 
The system has been implemented across two distinct embodied conversational agents (ECAs): ERICA \cite{glas2016erica, inoue2016talking, kawahara2019spoken}, a human-like android robot, and TELECO \cite{horikawa2023cybernetic}, a less anthropomorphic, teleoperated humanoid robot.

\subsection{Speech Processing}

For automatic speech recognition (ASR) and the extraction of prosodic features, we utilize a hand microphone. 
The ASR system is implemented via a real-time ASR module\footnote{\url{https://github.com/KoljaB/RealtimeSTT}}, which is based on the faster-whisper model\footnote{\url{https://github.com/SYSTRAN/faster-whisper}}. 
This setup facilitates the extraction of critical prosodic information, including fundamental frequency (F0) and power, from the spoken input.

\subsection{Dialogue Manager}

\begin{figure}[h]
  \centering
  \includegraphics[width=\linewidth]{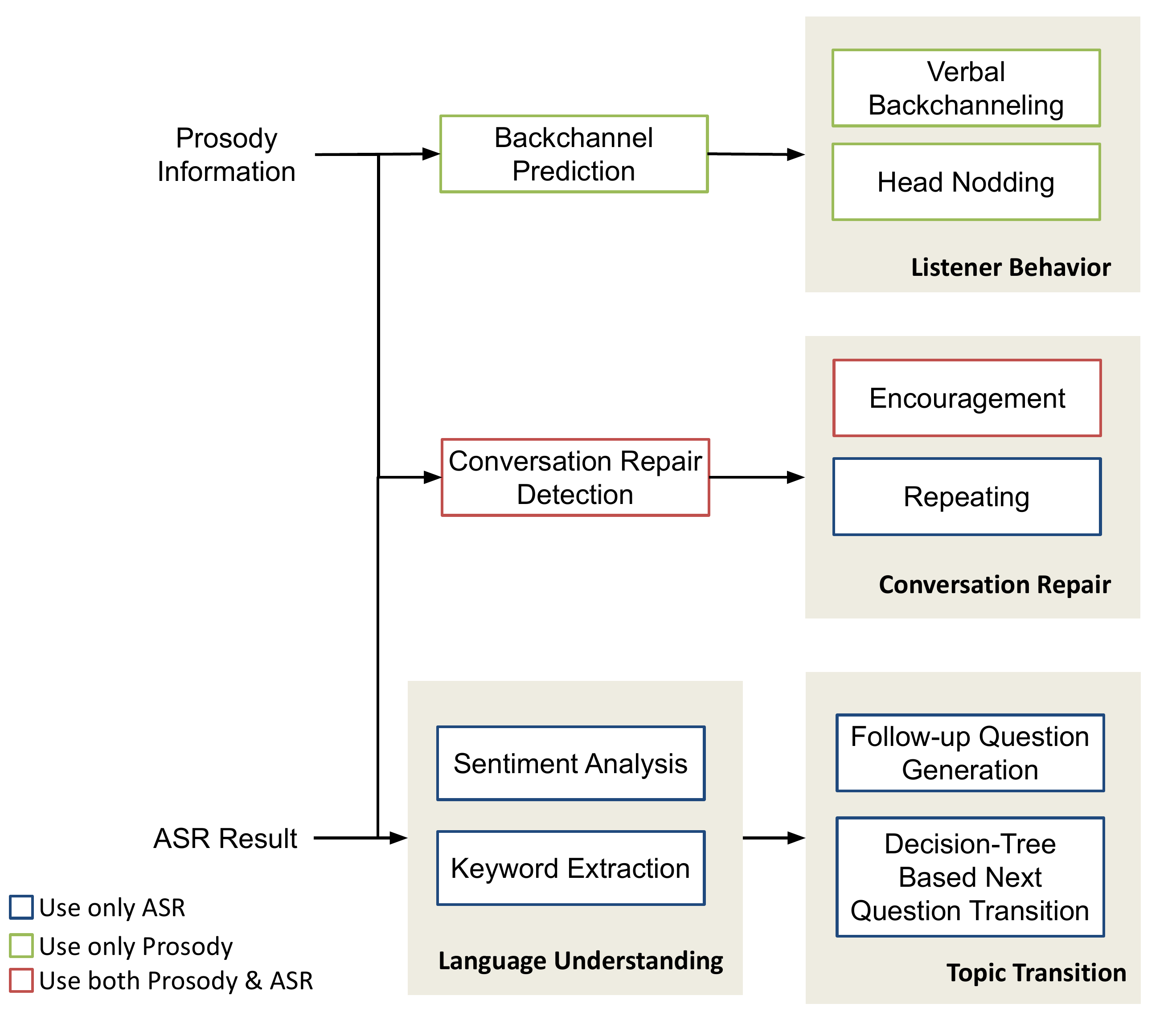} 
  \caption{Overall architecture of interview system response generation}
  \label{fig:response_architecture}
\end{figure}

The dialogue manager, a key component in our interview system, manages response selection based on user input. 
It comprises several sub-modules that improve interaction quality: a language understanding module that interprets user context to generate follow-up questions or smooth transitions; a backchannel module that predicts and delivers verbal and non-verbal cues, adding naturalness to the conversation; and a conversation repair module that detects and corrects communication breakdowns. 
Figure \ref{fig:response_architecture} depicts the architecture of the dialogue manager, highlighting the interplay among these components in response generation.

\subsubsection{Language Understanding}

The language understanding module uses ASR outputs for sentiment analysis, identifying keywords indicative of positive, neutral, and negative sentiments from a predefined polarity word list. 
For generating follow-up questions, the system considers context length and keyword presence—significant keywords include ``because'' and ``as.'' A follow-up question is generated if the context is under five words or lacks these keywords, using our predefined list of questions. 
Utilizing a decision-tree method, responses indicating agreement or disagreement guide the direction of subsequent questions. 
An example of dialogue processing and response generation by our system is detailed in the Appendix \ref{app:understanding dialogue}.

\subsubsection{Backchannel}

Backchanneling, where listeners indicate attentiveness through verbal, non-verbal, or combined responses, is crucial in conversations. 
Previous research has documented its use across languages \cite{cutrone2005case, ike2010backchannel} and settings \cite{widiyati2016features, maynard1986back}, including interviews \cite{wulandari2017conversation, nurjaleka2019backchannel,laforest1994listening}. 
Effective backchanneling and active listening enhance the interviewer's appeal and improve response quality \cite{louw2011active, rogers1957active, nurjaleka2023backchannels}. 
Despite advancements in LLMs, generating appropriate backchannels remains challenging, underscoring their importance in achieving human-like conversations. 
Previous human-robot interactions have primarily used non-verbal cues like head nodding without verbal responses \cite{inoue2021job}.

To address this, our system separates backchannel prediction and generation. 
We utilize a Multilingual-VAP based model \cite{inoue2024multilingual}, fine-tuned with attentive listening data, to predict appropriate moments for backchanneling based on prosodic cues. 
For generation, we developed a repertoire of verbal backchannels—such as ``hmm,'' ``erm,'' and ``mhmm''—and diverse head-nodding patterns that vary in frequency and speed for use in conversations. 
This approach supports simultaneous verbal and non-verbal backchannels to enhance the realism and effectiveness of the robot.

\subsubsection{Conversation Repair}

Conversation breakdowns frequently disrupt dialogues, particularly in spoken interviews, and can stem from issues from either the user or the system. 
For users, misunderstandings or difficulty in expressing thoughts can cause interruptions, whereas, for the system, challenges such as unrecognized speech or delays in processing can impede conversational flow.

To mitigate these disruptions, our system incorporates a conversation repair module that employs strategies of repeating and encouraging based on keyword detection. 
Utilizing prosodic cues and ASR results, the module identifies phrases indicating confusion, like ``pardon?'' or ``could you say that again?'' to repeat questions for clarity. 
Similarly, if expressions such as ``I have no idea'' or ``I don't know'' are detected, the system offers supportive responses, encouraging users to continue sharing their thoughts.

In cases where speech recognition fails despite clear voice activity, the system may use simple backchannels like ``mhmm'' to encourage continuation. 
Furthermore, to address processing delays that might lead to pauses, the system deploys interim responses like ``That's interesting!'' or ``That's a good point!'' from a predefined list, maintaining the conversational momentum while preparing the next question.

\begin{figure*}[t]
  \centering
  \includegraphics[width=13cm]{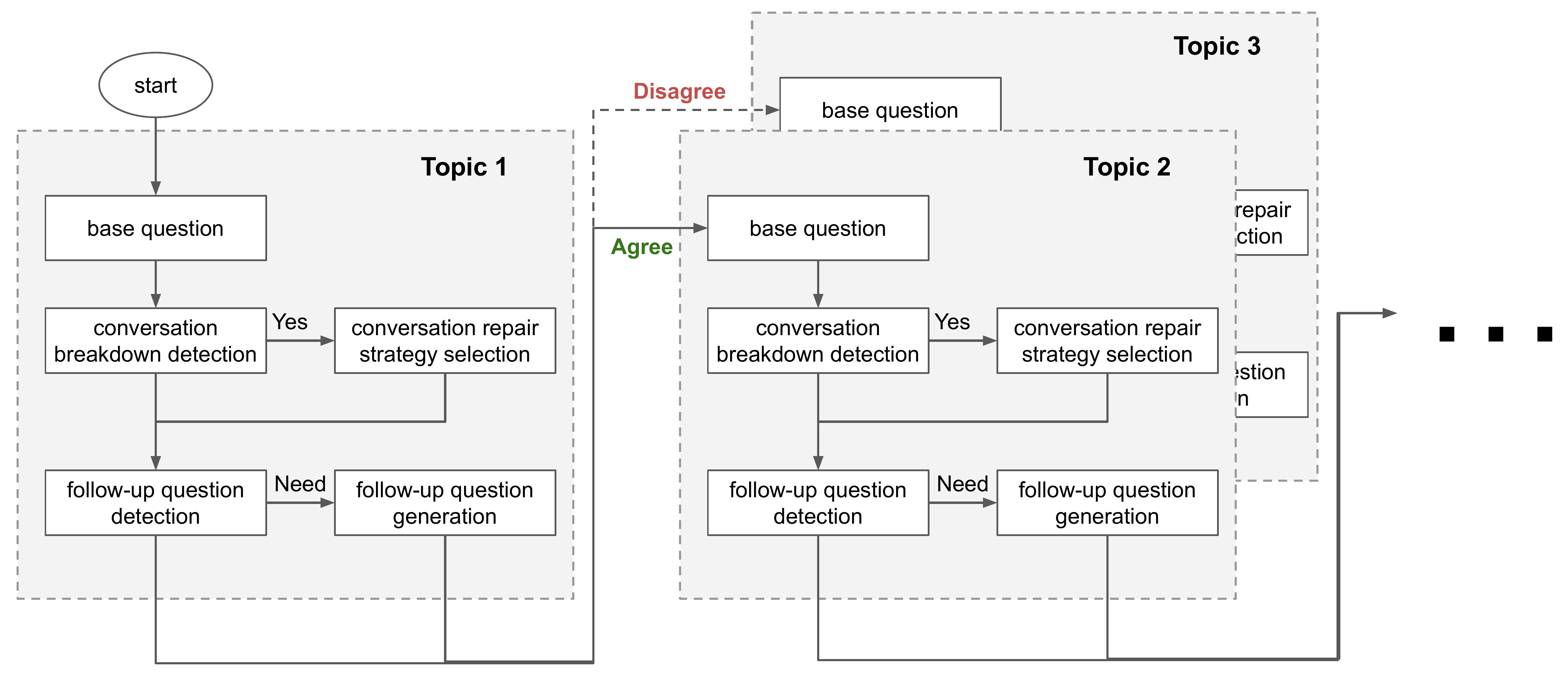} 
  \caption{Overall architecture of interview dialogue flow}
  \label{fig:dialogueflow}
\end{figure*} 

\subsubsection{User Fluency Adaptation}

User fluency significantly affects the smoothness of conversational flow during interviews. 
Fluent users usually engage without issues, but those less proficient may need additional time to articulate their thoughts, often resulting in longer silences and potential misunderstandings if the conversational pace is too rapid. 
This is particularly the case with non-native speakers. 
Our system includes a user fluency adaptation module that adjusts speaking speeds and extends turn-taking intervals according to user proficiency.

This module utilizes a Words-Per-Minute (WPM) based strategy, specifically designed to accommodate users with a WPM of 75 or below—indicative of beginner levels A1 to A2 according to the Common European Framework of References for Languages (CEFR)\footnote{\url{https://magoosh.com/english-speaking/english-proficiency-levels-a-guide-to-determining-your-level/}}. 
For these users, the system slows down its speech and allows longer response times. 
This adaptation helps non-fluent speakers engage effectively with our system, as standard conversational speeds in English, typically between 150-190 WPM and reaching up to 197 WPM in formal interviews \cite{marslen1973linguistic, richards1983listening, wang2021british}, far exceed what beginners can handle. 
Even academic presentations, which generally maintain a slower pace of 100–125 WPM for clarity \cite{wong2009essential}, surpass optimal speeds for these users. 
Utilizing user fluency adaptation, our system ensures the interview process accommodates speakers of varying proficiency, which is crucial in international conferences with diverse linguistic backgrounds.

\subsection{Speech Synthesis}
\label{sec:speech synthesis}

For speech synthesis, our system uses the Julie voice provided by the VoiceText engine from Hoya Corporation\footnote{\url{https://readspeaker.jp/}}. 
Although this engine capably synthesizes standard speech, it struggles with the nuanced pronunciation of verbal backchannels such as ``mhmm'' or ``hmm''. 
These elements are crucial for natural conversational flow but are not adequately represented when generated directly by typical text-to-speech (TTS) systems due to their unique phonetic characteristics.

To overcome this limitation, we manually adjusted and refined the pronunciation of each backchannel, subsequently creating the corresponding .wav files. 
This approach allows our system to incorporate a diverse array of backchannels, varying in form and speed, to enhance the realism and dynamic nature of interactions.

\subsection{Gesture Generation}

To enhance the human-like quality of our interview system, we developed a range of gestures that extend beyond mere head nodding. 
Among these, an open palm gesture, which signifies openness and accessibility, fostering an environment conducive to free expression and interaction\footnote{\url{https://www.globallisteningcentre.org/body-language-of-listeners/}}. 
Additionally, we have implemented gestures such as leaning back to indicate surprise during interactions, and a bowing gesture to signify respect and formality after completing an interview. 
These gestures are strategically designed to mimic human non-verbal cues, thereby enhancing the naturalness and effectiveness of the robot's interactions with users.

\subsection{Interview Dialogue Flow}

\begin{figure*}[t]
  \centering
  \includegraphics[width=14cm]{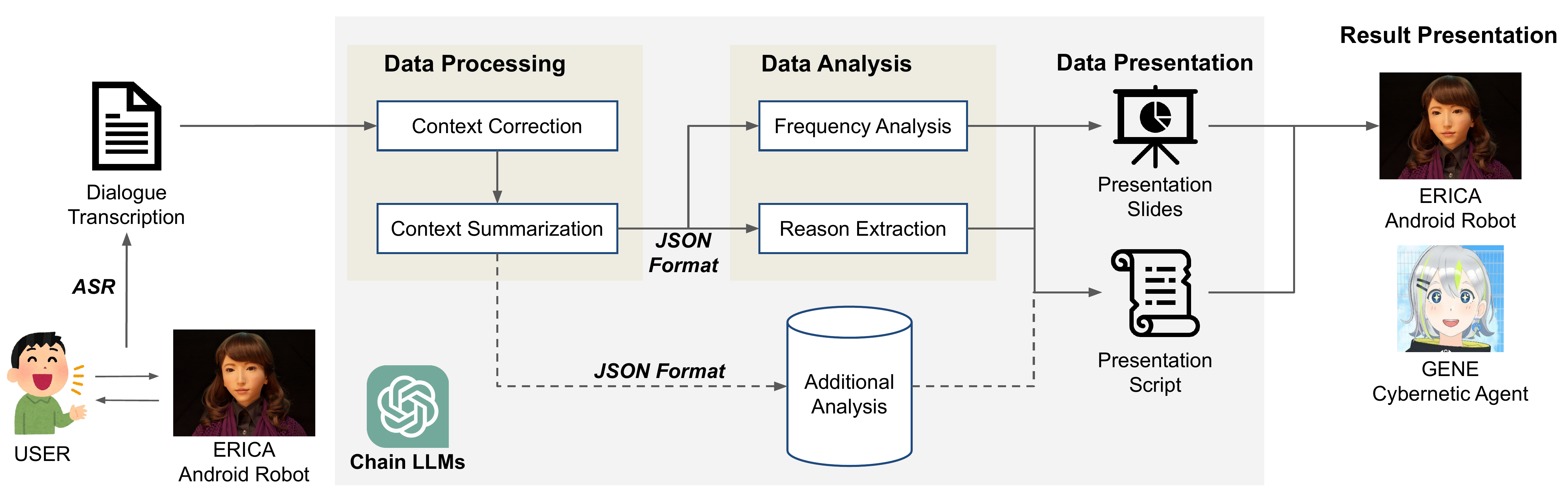} 
  \caption{Overall architecture of post-interview processing workflow}
  \label{fig:workflow}
\end{figure*}

The interview dialogue flow in our human-like interview system is managed via finite state transitions, as depicted in Figure \ref{fig:dialogueflow}. 
The process initiates with a base question. 
Based on the user's response, the system evaluates whether a conversational breakdown has occurred and if interventions, such as repeating or encouraging, are necessary. 
If such responses are required, the system generates them and maintains the current question state. 
Otherwise, the system assesses whether the user has provided sufficient information. 
If the information is inadequate, a follow-up question is posed. 
Subsequently, the system determines the next set of questions to be addressed based on the user's latest response. 
This cycle continues throughout the interview. Concurrently, to enhance human likeness and express interest in the user's responses, the system delivers both verbal and non-verbal backchanneling while receiving user input.

\subsection{Interview Question Strategy}
For the interview question strategy, we adopted a hybrid approach that integrates both template-based and generative question sets. On one hand, we employed LLMs (i.e., GPT-4o-mini API\footnote{\url{https://openai.com/index/gpt-4o-mini-advancing-cost-efficient-intelligence/}}) to dynamically produce follow-up inquiries, thereby adapting naturally to user responses and maintaining a human-like conversational flow. On the other hand, we use a fixed set of template-based questions for the primary prompts central to our data collection, ensuring that these core questions remain consistent across all interviews. This balance not only supports reliable downstream analysis but also enables adaptability through the generative component. Additionally, the system’s modular design allows for flexible expansion for both the template-based and generative prompts, reducing the need for extensive manual rule-crafting. However, to ensure the highest level of analytic rigor in a research setting, the real-world case study presented in Section \ref{sec:case studies} relied solely on the template-based approach, maintaining question stability, which is necessary for accurate evaluation.

\subsection{Post-Interview Processing Workflow}

The post-interview processing workflow in our system facilitates data processing, analysis, and presentation. 
Utilizing a series of chained LLMs, specifically GPT-4o-mini\footnote{\url{https://openai.com/index/gpt-4o-mini-advancing-cost-efficient-intelligence/}}, our system segments tasks into distinct subtasks with targeted prompts. 
This modular approach enhances task specificity and enables precise control over the process, allowing for modifications at any stage to suit specific research needs. 
The workflow's structure is detailed in Figure \ref{fig:workflow}.

The pipeline consists of three main phases: data processing, analysis, and presentation. 
Initial data processing corrects ASR errors, ensuring data integrity, and prepares data in JSON format for subsequent analysis. 
The analysis involves evaluating opinion distributions and motivations, with flexibility for additional inquiries. 
Presentation materials, such as scripts and slides, are generated from the analysis results, using tools like the python-pptx library\footnote{\url{https://python-pptx.readthedocs.io/en/}}. 
This automated system concludes with presentations delivered by conversational agents such as robots or virtual agents. 
Each subtask's detailed prompts are provided in the Appendix \ref{app:prompt}.

\section{Real-World Case Study}
\label{sec:case studies}

To evaluate our human-like embodied AI interviewer's effectiveness in a real-world setting, we conducted a case study at SIGDIAL 2024, attended by over 160 participants\footnote{\url{https://2024.sigdial.org/}}. 
This study assessed perceptions of conversational AI's human-likeness, exploring themes such as essential interaction qualities, the importance of human-like traits, the inclusion of negative traits, and strategies against misuse.

\begin{figure}
  \centering
  \includegraphics[width=\linewidth]{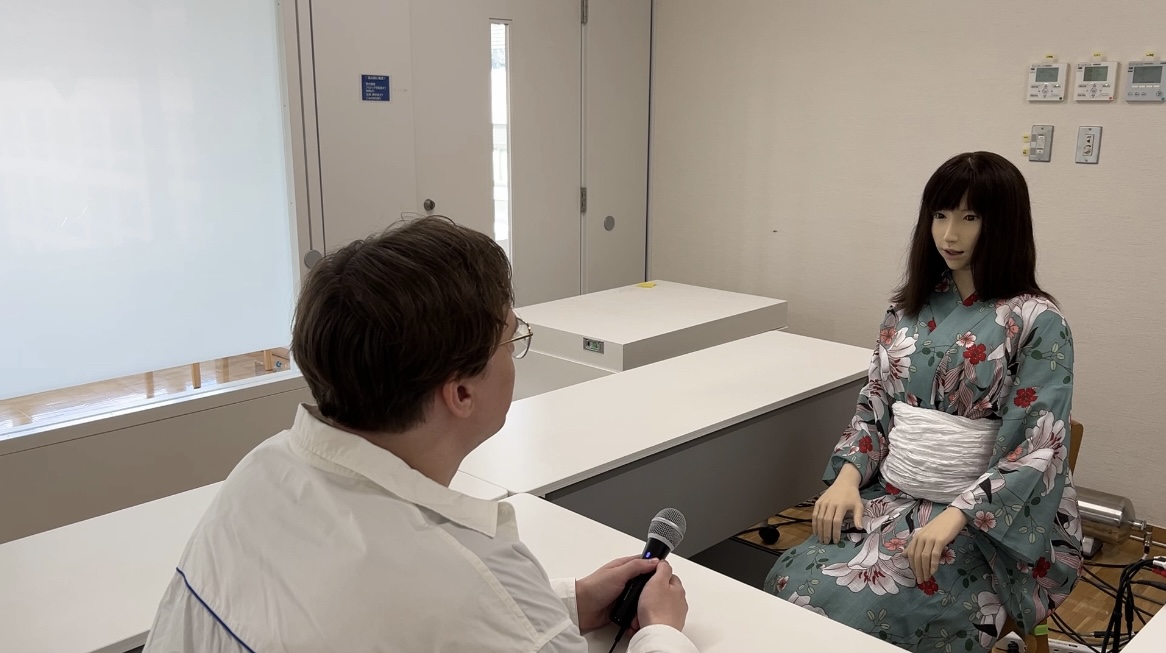} 
  \caption{
  Photo of interview dialogue with ERICA by SIGDIAL participant
  }
  \label{fig:erica}
\end{figure}

Participants engaged in brief interviews lasting 2-3 minutes with one of two robots at the conference: ERICA, an android resembling a female adult, and TELECO, a humanoid robot with an OLED display face and simplified joint structures. 
Both robots exhibited identical dialogue behaviors, gestures, and facial expressions. 
Figure \ref{fig:erica} illustrates a user interacting with ERICA during the conference\footnote{Demo video is available at \url{https://youtu.be/v1vfRJu_UJ4}}. Another user interacting with TELECO during the conference is illustrated in Figure \ref{fig:teleco}.
The example dialogues are provided in Appendix \ref{app:example}, and the setup details and interview results are documented in Appendix \ref{app:result}.

Interviews occurred over the first two days of the conference, with findings presented on the final day. To ensure a natural interaction environment, no formal questionnaire feedback was solicited. 
Instead, experiences were gathered directly during the interviews and through spontaneous post-interview discussions with participants. 
Insights from these interactions are elaborated in the subsequent subsection.

Consent was obtained by informing attendees at the conference's opening session and through clearly displayed notices in the interview room, advising that only transcripted dialogues from ASR would be recorded.

\begin{figure}
  \centering
  \includegraphics[width=\linewidth]{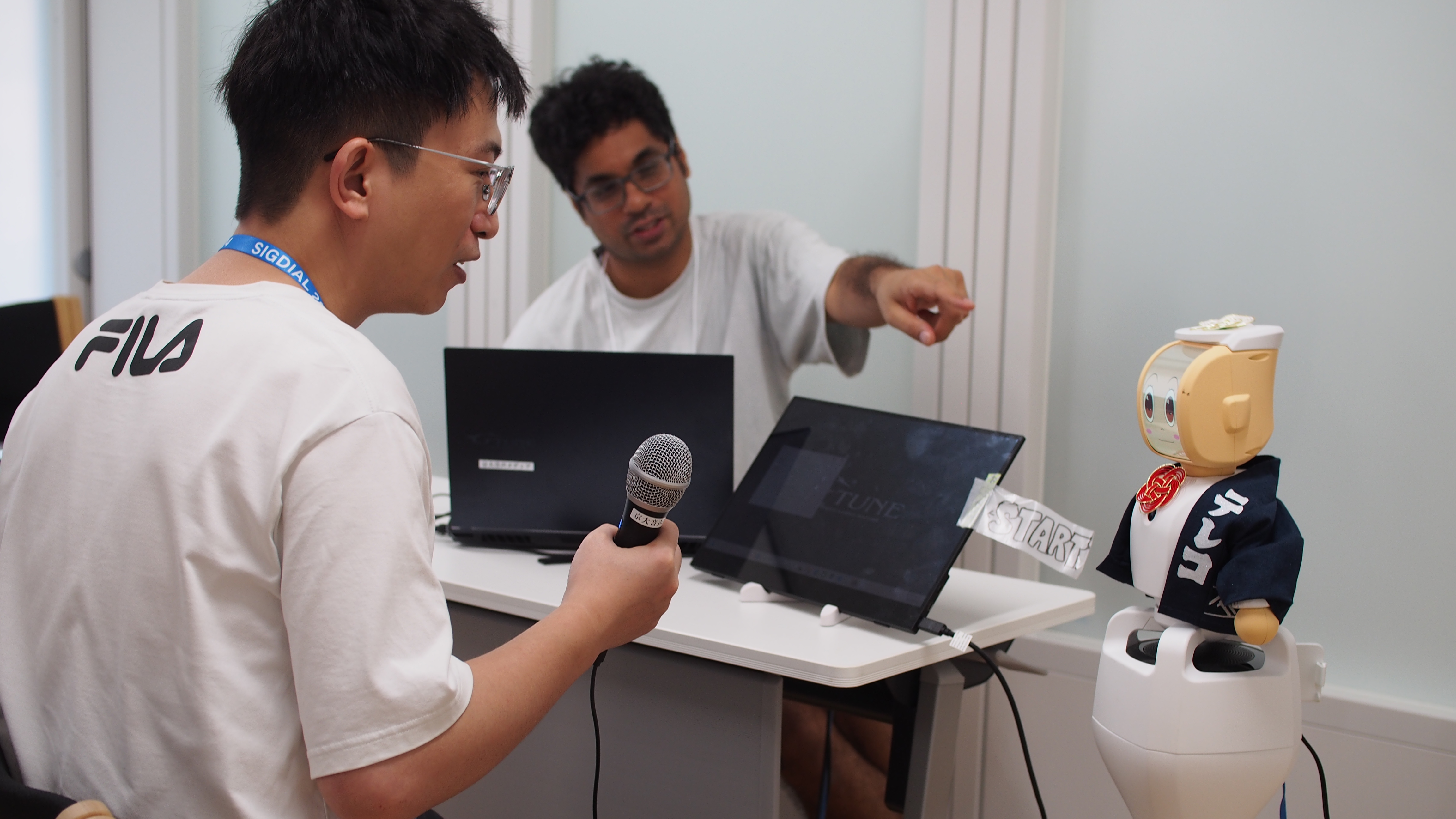} 
  \caption{Photo of interview dialogue with TELECO by SIGDIAL participant}
  \label{fig:teleco}
\end{figure}

\subsection{Reporting on Panel Discussion}
In academia, panel discussions usually involve a group of experts and a moderator to foster an informative exchanges of viewpoints. 
Such settings are advantageous for gathering expert opinions across various fields, providing a deep understanding of specific topics \cite{rasmussen2008panel, tempero2011pancreatic, filbeck2017behavioral}. 
However, these discussions often face time constraints and typically limit participation to high-level experts like professors, restricting the diversity of perspectives.

\begin{figure}
  \centering
  \includegraphics[width=\linewidth]{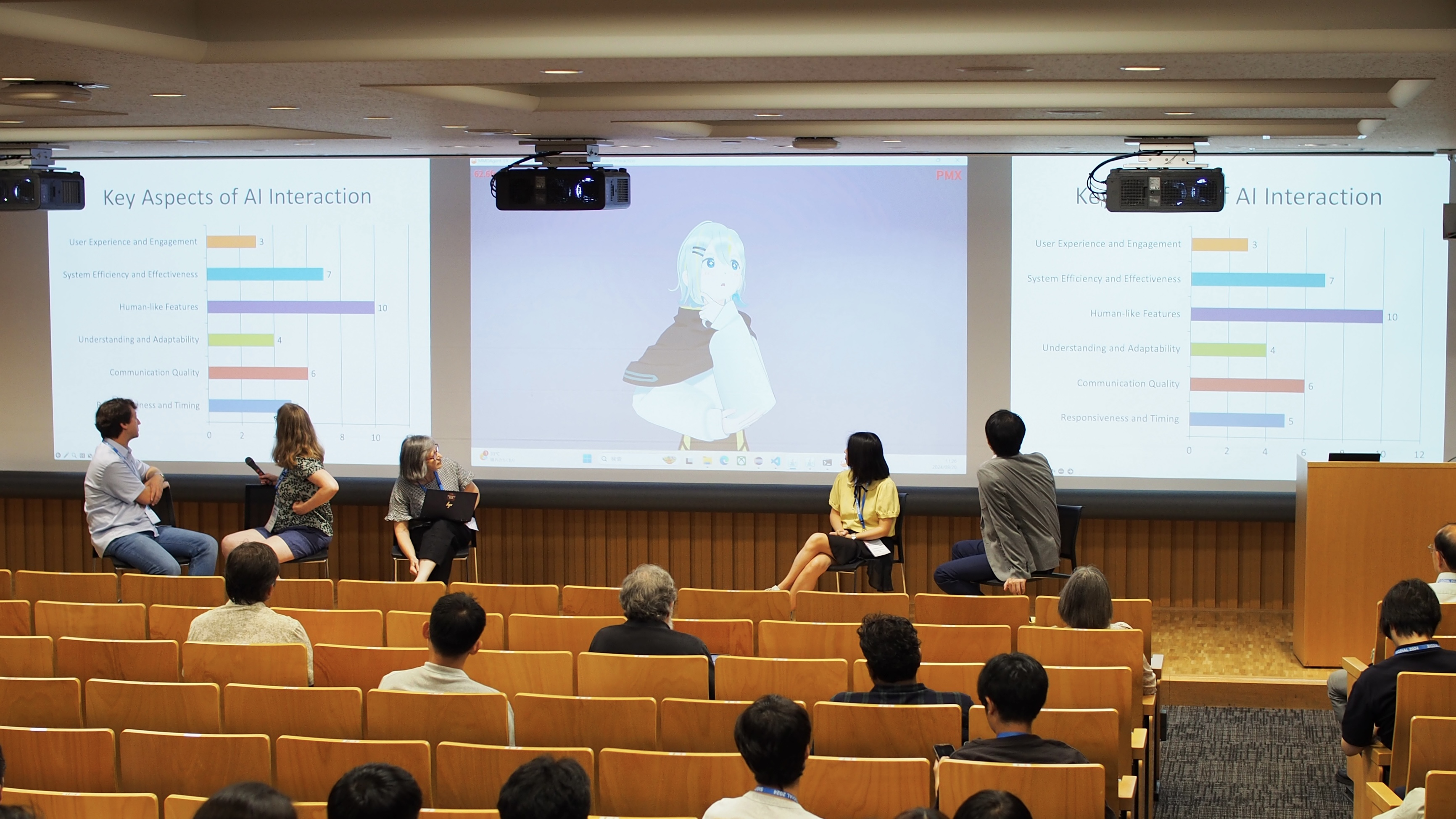} 
  \caption{Photo of Gene's presentation during the panel discussion session at SIGDIAL}
  \label{fig:gene}
\end{figure}

Our system extends beyond this limitation by collecting opinions from conference participants at all levels, not just from high-level experts, ensuring that all participants had the opportunity to express their opinions. During the panel discussion session on the last day, our system presented the analyzed results. 
Due to logistical challenges, instead of presenting with ERICA on stage, we utilized a computer-generated (CG) agent, Gene \cite{Lee_MMDAgent-EX_2023}, who presented the results\footnote{Demo video is available at \url{https://youtu.be/pSgaouAUkZk}.}. The presentation by Gene is illustrated in Figure \ref{fig:gene}.

\subsection{Result and Discussion}

The feedback from participants was predominantly positive, affirming the system's effectiveness in facilitating engaging and memorable interactions. 
Of 42 participants, 29 described the interaction as ``enjoyable and engaging'' and felt it encouraged them to share more thoughts. 
These comments reflect the system’s success in engaging users effectively, with the overall results illustrated in Table \ref{tab:result}.

However, not all feedback was positive. 
From two participants, critical insights emerged, highlighting the repetitive nature of the interview, with remarks like ``The interview felt repetitive as the robot asked fixed questions.'' 
This feedback underscores the need for more adaptive and personalized follow-up questions, potentially through enhanced use of LLMs to enable dynamic conversation flows. 
Additionally, some participants expressed discomfort with the robots’ human-like appearance, indicating a need for careful calibration to balance human-likeness and user comfort.

Further feedback from spontaneous post-interview conversations revealed mixed reactions to the system’s backchanneling capabilities. 
While many appreciated the verbal and non-verbal backchannels for enhancing the perception of attentiveness and human-likeness, there were criticisms about the naturalness of synthesized backchannels like ``mhmm,'' suggesting that the current speech synthesis engine may not effectively capture the casual tone required for everyday conversational backchannels. 
This opens avenues for future research into developing more human-like and context-appropriate backchannel generation.

Discussions also revealed diverse preferences concerning the appearances of our robots, particularly between the highly human-like android ERICA and the less human-like humanoid TELECO. 
Some participants found ERICA’s resemblance unsettling, while others valued the genuine sense of co-presence she provided. 
In contrast, TELECO’s less human-like features did not evoke the same level of co-presence. 
These varied responses highlight cultural or personal differences in acceptance and preference of robot aesthetics, suggesting a rich area for further investigation into how culture and personality influence human reactions to the human-likeness of robots.

\begin{table}[h]
\caption{Overall Interview Experience Result [\%]}
\label{tab:result}
\centering
{\small
\begin{tabular}{lp{52mm}}
\hline
\multicolumn{1}{c}{Experience} & \multicolumn{1}{c}{Common Reasons} \\
\hline
\rowcolor[gray]{0.95} Positive & 1. Interaction is engaging \\
\rowcolor[gray]{0.95} \textbf{(69.05)} & 2. Interesting human-like robot \\

\multirow{2}{*}{\shortstack{Neutral\\(26.19)}} & 1. Interesting but experienced an error \\
 & 2. Interesting but wanted more support \\

\rowcolor[gray]{0.95} Negative & 1. Questions felt repetitive \\
\rowcolor[gray]{0.95} (4.76) & 2. Robot appearance caused discomfort \\
\hline
\end{tabular}
}
\end{table}






\section{Conclusion}

In this paper, we introduced the human-like embodied AI interviewer, integrating android and humanoid robots with chained LLMs to support researchers in data collection, analysis, and presentation.
Our system improved interview quality by incorporating advanced conversational behaviors such as attentive listening, conversational repairs, and user fluency adaptation, and automated the analysis and presentation processes post-interview.

A two-day case study at an international academic conference validated our system's effectiveness, with 69\% of participants reporting positive experiences.
The system also streamlined data analysis and presentation.
Notably, this was the first use of such a system at an international conference, demonstrating its applicability in real-world research settings.

Looking forward, we aim to enhance the human-like features of our system, focusing on improving backchannel generation and exploring cultural and personal preferences for robot appearances to optimize user interactions.
We hope that these enhancements will bring us closer to achieving human-level interaction capabilities in android robots, further bridging the gap between technology and natural human communication.

\section{Limitations}

While our preliminary two-day case study at an international academic conference offered initial validation, the relatively small sample size (42 participants) limits the generalizability of our findings. To improve the generalizability of our findings, we plan to conduct larger-scale studies with more diverse participant groups.

Another limitation lies in the repetitive nature of the template-based questioning utilized in the real-world case study. While these fixed templates ensured stability for analysis, they reduced conversational variability. In future work, we aim to incorporate LLMs to generate questions more dynamically, exploring methods to maintain question stability without compromising adaptability and user engagement.

Lastly, the current system relies solely on speech input, which constrains its capacity to fully interpret users’ states and behaviors. Moving forward, we will integrate richer multimodal inputs—including facial expressions, body language, and environmental context—to achieve more responsive, context-aware interactions and enhance the overall user experience.

\section*{Acknowledgments}


The authors thank the participants of SIGDIAL 2024 for their contributions, which were crucial for the analysis and conclusions of this study. 
This research was supported by the JST Moonshot R\&D Goal 1 Avatar Symbiotic Society Project (JPMJMS2011).



\bibliography{custom}

\clearpage

\appendix

\section*{Appendix}

\appendix

\section{Language Understanding Dialogue Example}
\label{app:understanding dialogue}

Below is a dialogue example demonstrating our language understanding system. \textbf{R} indicates an utterance by the system, while \textbf{U} indicates an utterance by the user. Dialogue 1 demonstrates a scenario where the system generates a follow-up question due to the absence of keywords like ``because'' or ``as''. In contrast, Dialogue 2 shows a scenario where the system does not generate a follow-up question because the user spoke extensively. \newline

\noindent
\textbf{Dialogue 1} \\
\textbf{R1} What do you think is the most important thing when interacting with a conversational AI? \\
\textbf{U1} Uh, well, I would say the response time maybe. \\
\textbf{R2} Interesting! Can you tell me more about why you think that's so important? (Generated a follow-up question as no keywords were detected) \\

\noindent
\textbf{Dialogue 2} \\
\textbf{R1} What do you think is the most important thing when interacting with a conversational AI? \\
\textbf{U1} I think it should be that I am properly understood and my questions are addressed accurately. I think that if I'm talking with somebody, I really would like to be understood, so I think that's very important. \\
\textbf{R2} That's a good point! So, let’s move on. (Moved to the next question since the user spoke a lot) \\

The example dialogues below showcase how responses guide the flow of the interview, especially when determining the next question. In the \textit{importance of human-like traits} question, when keywords like ``agree'', ``yes'', ``yeah'' are detected indicating agreement, the next question shifts to \textit{the inclusion of negative traits}, as shown in Dialogue 3. Conversely, if the user disagrees, the next questions will be about \textit{strategies against misuse}, as shown in Dialogue 4. \newline

\noindent
\textbf{Dialogue 3} \\
\textbf{R1} Do you think conversational AI should aim to be human-like? \\
\textbf{U1} Yeah, not only human-like but also considering the user's preferences. \\
\textbf{R2} Okay, that's interesting! And what about negative human traits? Should conversational AI include these as well? (Moved to the \textit{inclusion of negative traits} topic since the user agreed) \\

\noindent
\textbf{Dialogue 4} \\
\textbf{R1} Do you think conversational AI should aim to be human-like? \\
\textbf{U1} Um, not really. I think that conversational AI can be useful even if it's not human-like. \\
\textbf{R2} Makes sense! So, what would you do to prevent misuse of conversational AI? (Moved to the \textit{strategies against misuse} topic as the user disagreed) \\

Sentiment analysis was implemented in our proposed system to determine the appropriate response. For instance, when querying user experiences about being interviewed by a robot, the robot's response will change based on the sentiment of the user's utterance. This adaptation is illustrated below in Dialogue 5 and Dialogue 6.

\noindent
\textbf{Dialogue 5} \\
\textbf{R1} How did you feel about being interviewed by me, a human-like robot?  \\
\textbf{U1} Yeah, it was a really interesting experience because this is my first time. \\
\textbf{R2} I'm glad that you enjoyed this conversation. I appreciate your time! Have a wonderful day!  (Concludes positively in response to the user's enjoyment) \\

\noindent
\textbf{Dialogue 6} \\
\textbf{R1} How did you feel about being interviewed by me, a human-like robot?\\
\textbf{U1} It's a little creepy. \\
\textbf{R2} I'm sorry to hear that, but I appreciate your honesty. Thank you for your time. (Responds courteously to the user’s negative feedback) \\

\section{Prompts}
\label{app:prompt}

\begin{figure*}[h]
  \centering
  \includegraphics[width=\linewidth]{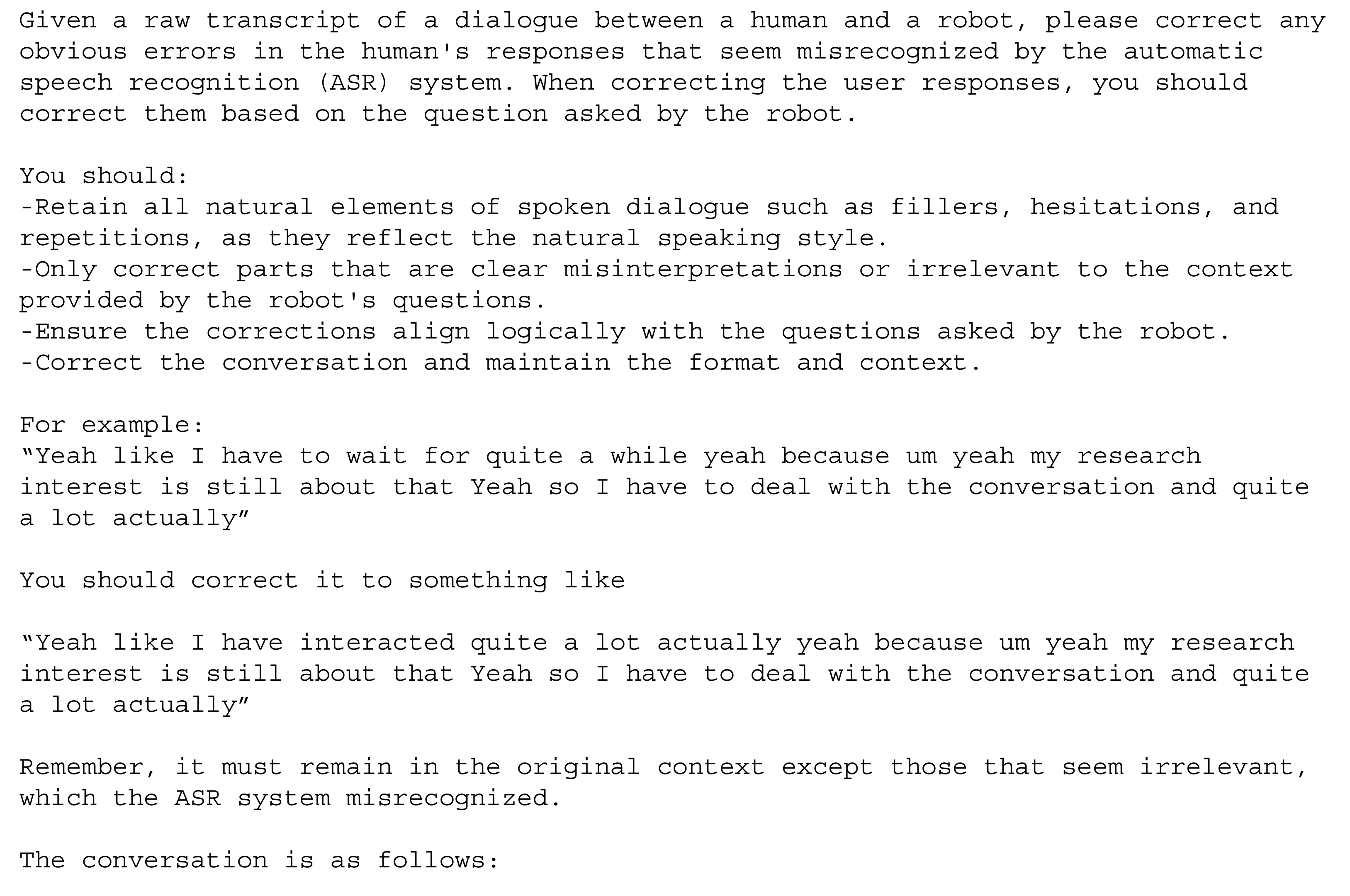} 
  \caption{Prompt for correcting dialogue context due to ASR error}
  \label{fig:correct}
\end{figure*}

\begin{figure*}[h]
  \centering
  \includegraphics[width=\linewidth]{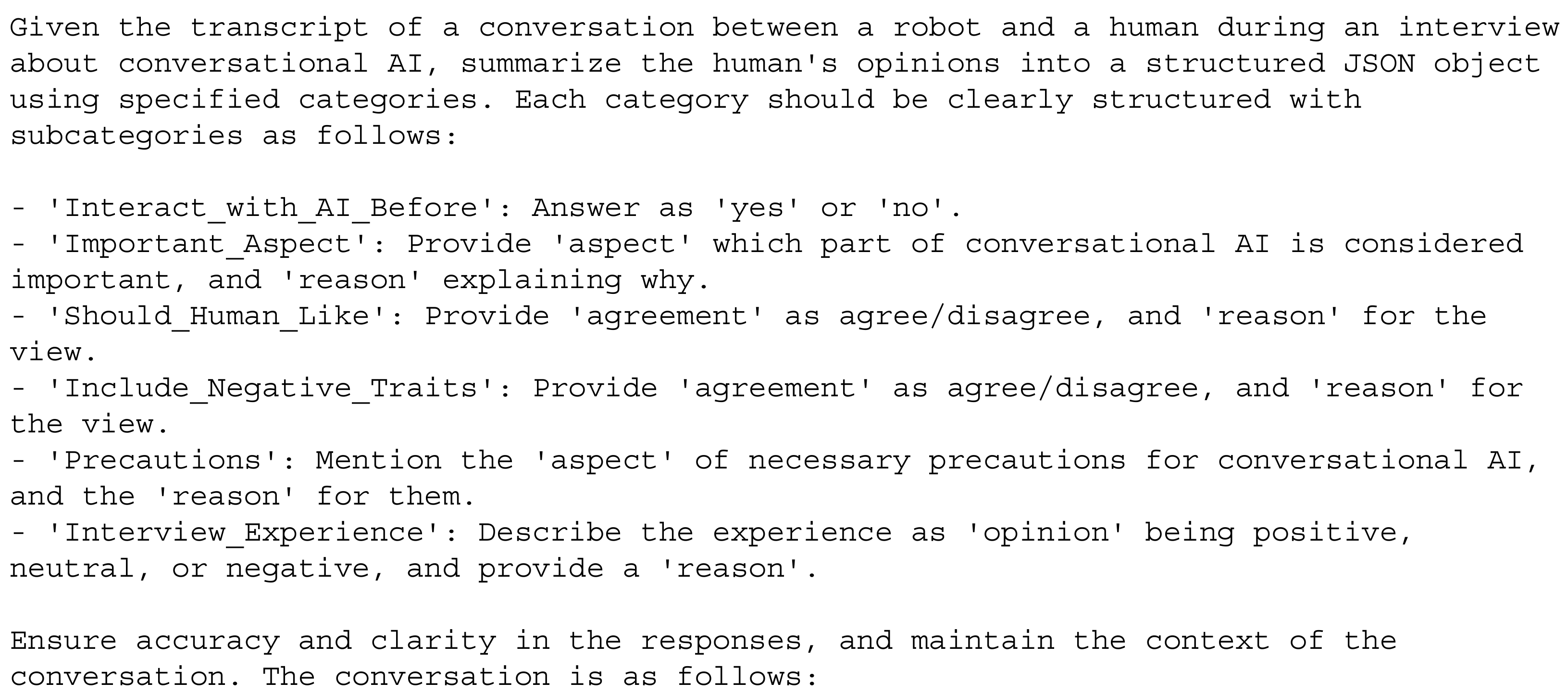} 
  \caption{Prompt for summarizing dialogue context into JSON format}
  \label{fig:summarize}
\end{figure*}

\begin{figure*}[h]
  \centering
  \includegraphics[width=\linewidth]{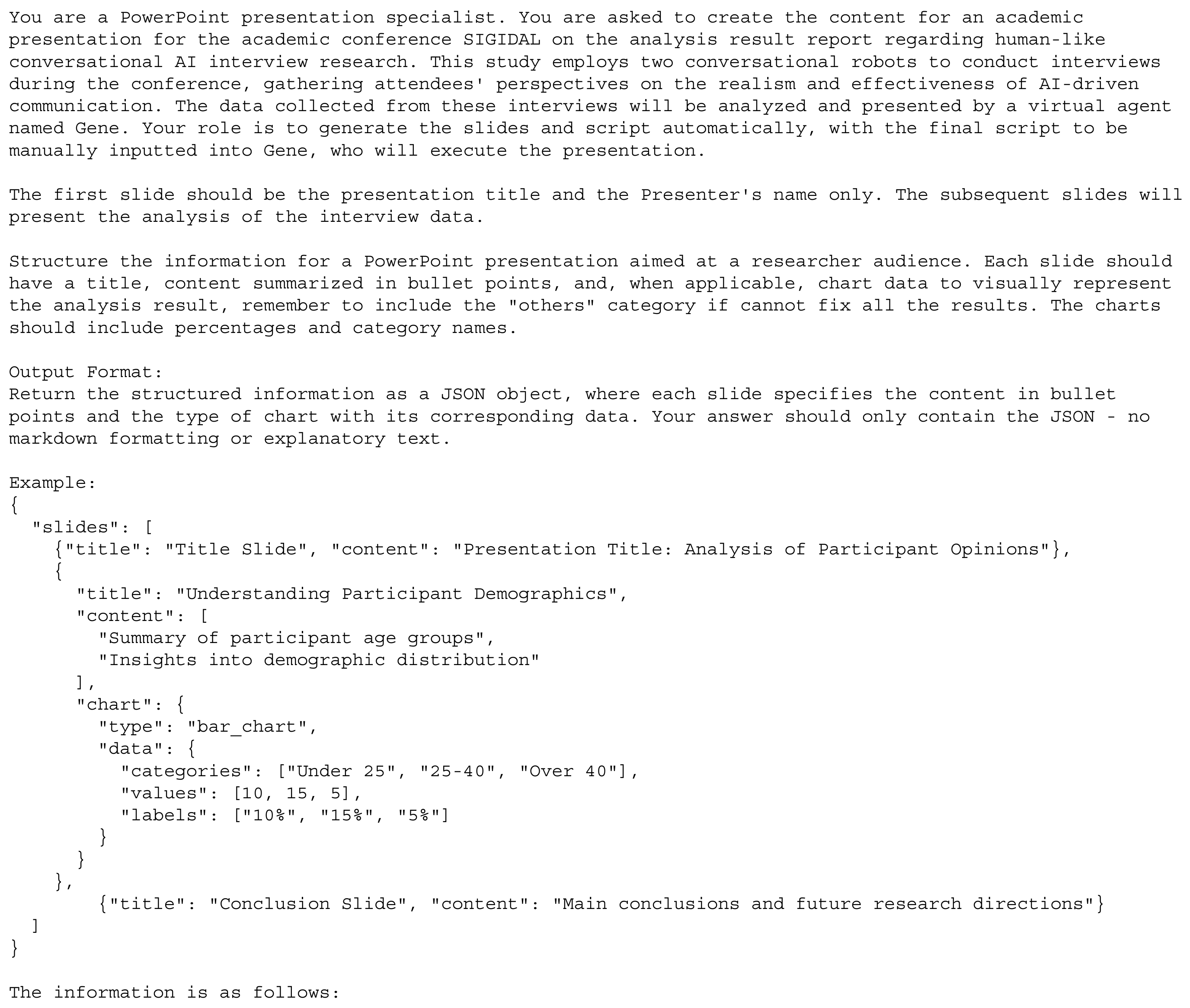} 
  \caption{Prompt for generating presentation slide context}
  \label{fig:slide}
\end{figure*}

\begin{figure*}[h]
  \centering
  \includegraphics[width=\linewidth]{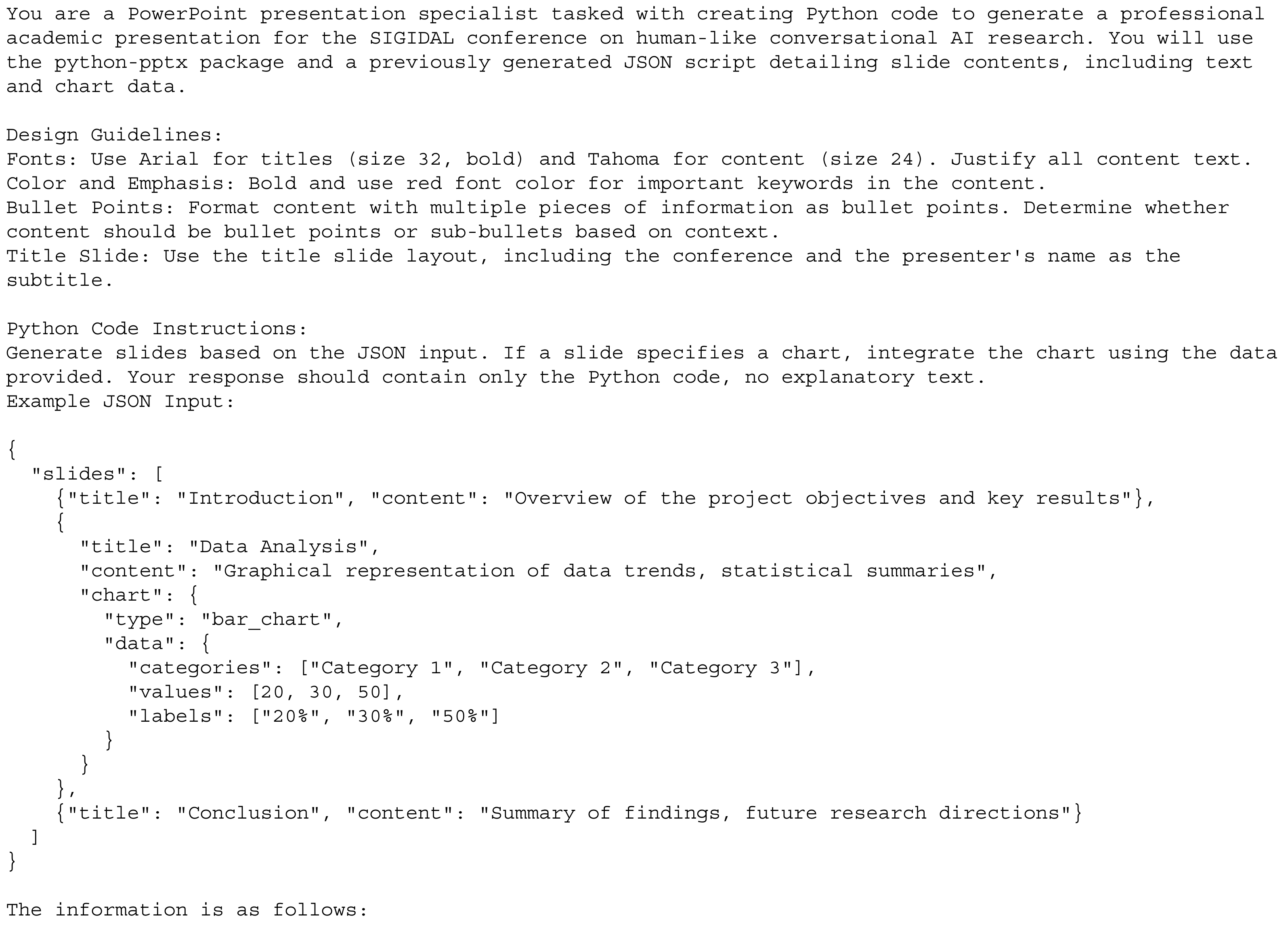} 
  \caption{Prompt for generating presentation slide python script from the presentation slide context}
  \label{fig:slidecode}
\end{figure*}

\begin{figure*}[h]
  \centering
  \includegraphics[width=\linewidth]{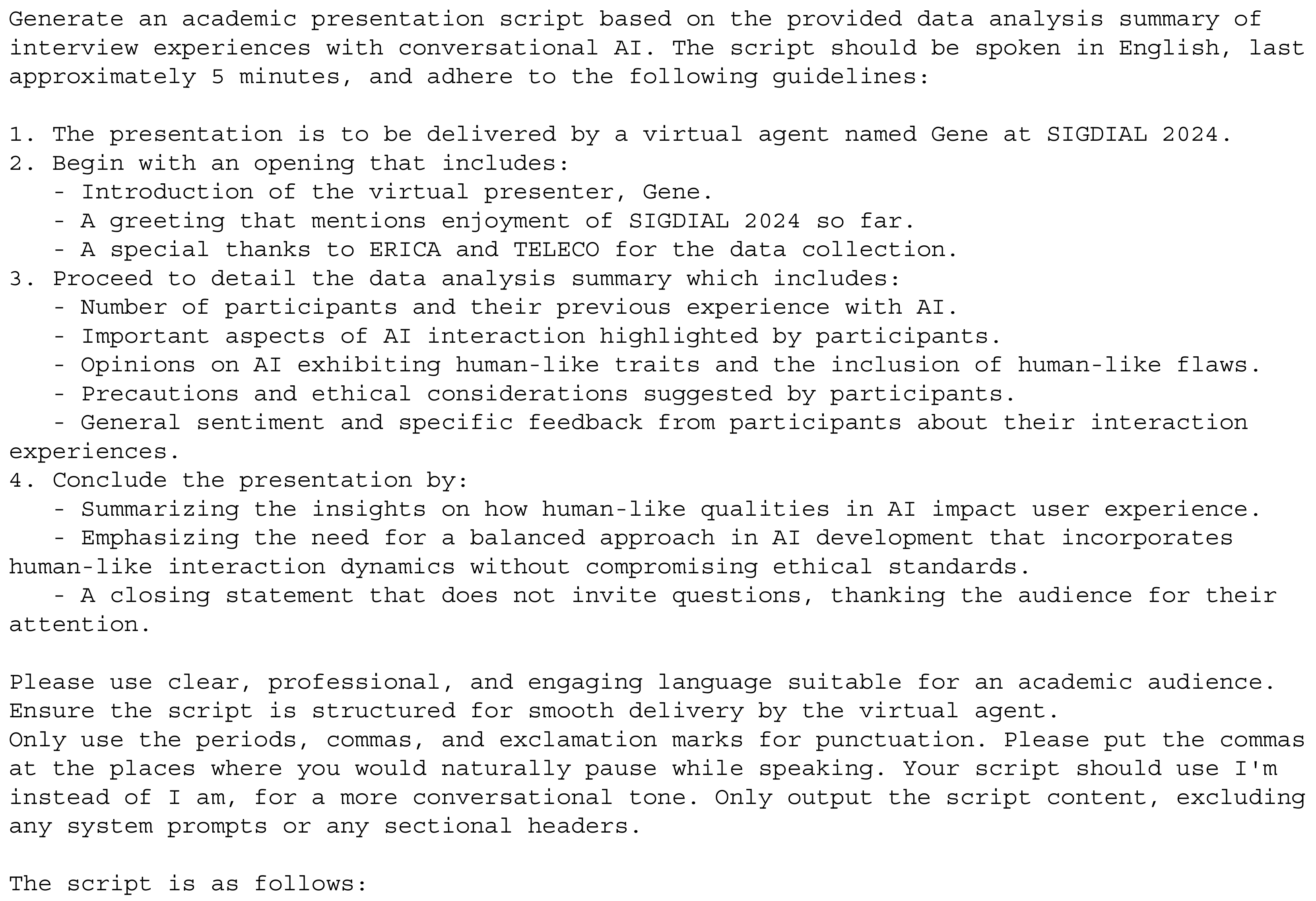} 
  \caption{Prompt for generation presentation script}
  \label{fig:scriptcontext}
\end{figure*}

Below are the prompts provided to the LLMs to facilitate response generation. 
In the post-interview processing workflow, our system used a series of chained LLMs to handle data through context correction, summarization, data analysis, and the generation of presentation slides and scripts. 
The pipeline involves a cascading approach where the input from each subtask is passed to the next task. 
This approach comprehensively manages processing, analysis, and presentation generation. Detailed prompts for each subtask are depicted in \cref{fig:correct,fig:summarize,fig:slide,fig:slidecode,fig:scriptcontext}.

\section{Case Study Dialogue Example}
\label{app:example}

\begin{figure*}[h]
  \centering
  \includegraphics[width=\linewidth]{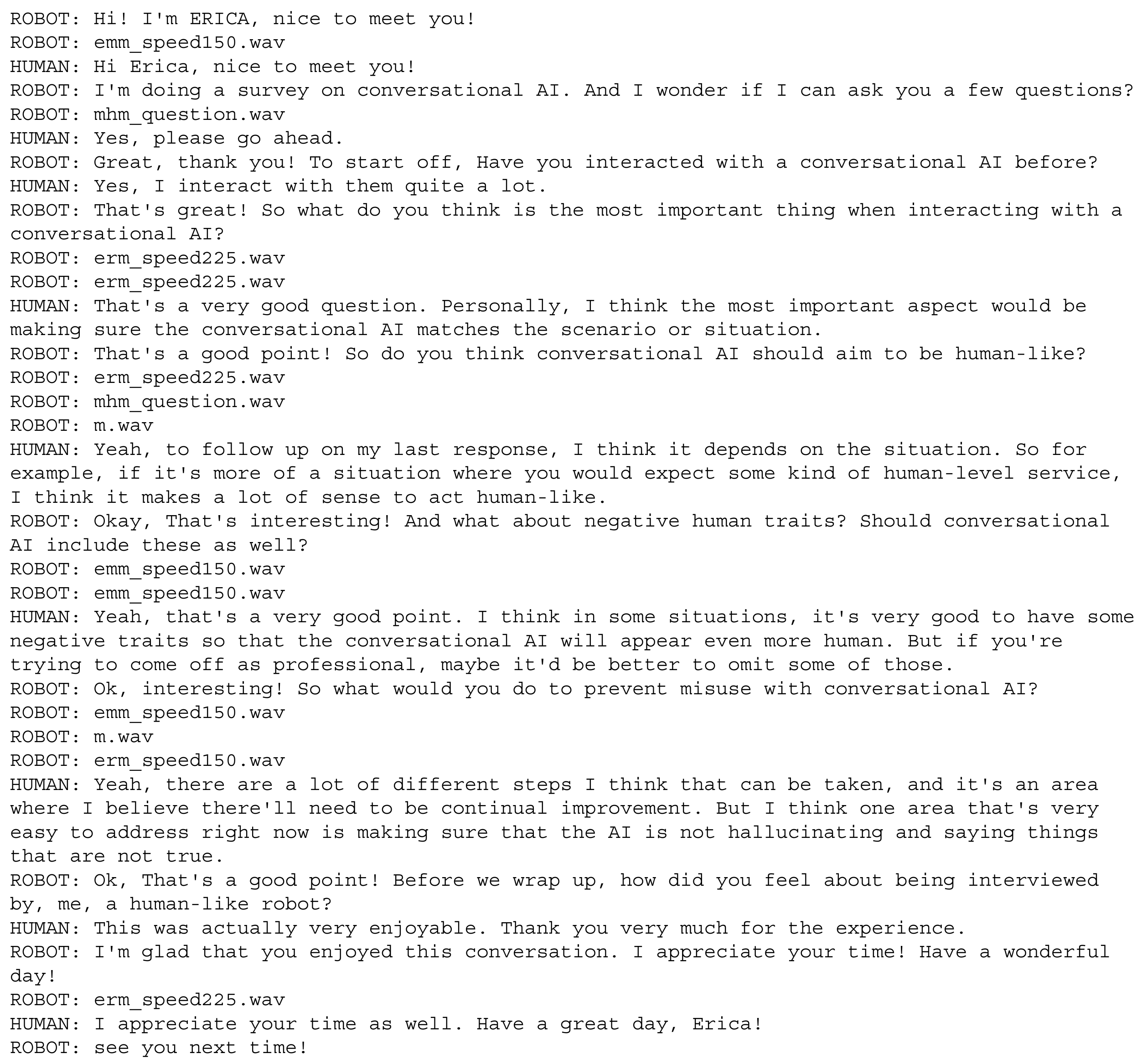} 
  \caption{Dialogue example collected during the case studies}
  \label{fig:example}
\end{figure*}

See Figure \ref{fig:example} for a dialogue example\footnote{Demo video is available at \url{https://youtu.be/v1vfRJu_UJ4}} that explores participant perceptions of the \textit{human-likeness of conversational AI}. 
This example addresses four primary topics: essential interaction qualities, the importance of human-like traits, the inclusion of negative traits, and strategies to prevent misuse. In the dialogue, \textbf{ROBOT} denotes system utterances, while \textbf{HUMAN} represents user responses. 
As detailed in Section \ref{sec:speech synthesis}, verbal backchannels cannot be directly synthesized by our speech engine; therefore, we manually created these sounds and played the corresponding .wav files as needed. 
In the dialogue example, any system utterance ending with .wav indicates a generated verbal backchannel. 
Due to the simultaneous occurrence of verbal backchannels and user utterances, the log file records system backchannels before user responses.

\section{Case Study Details}
\label{app:result}

\begin{figure*}[h]
  \centering
  \includegraphics[width=\linewidth]{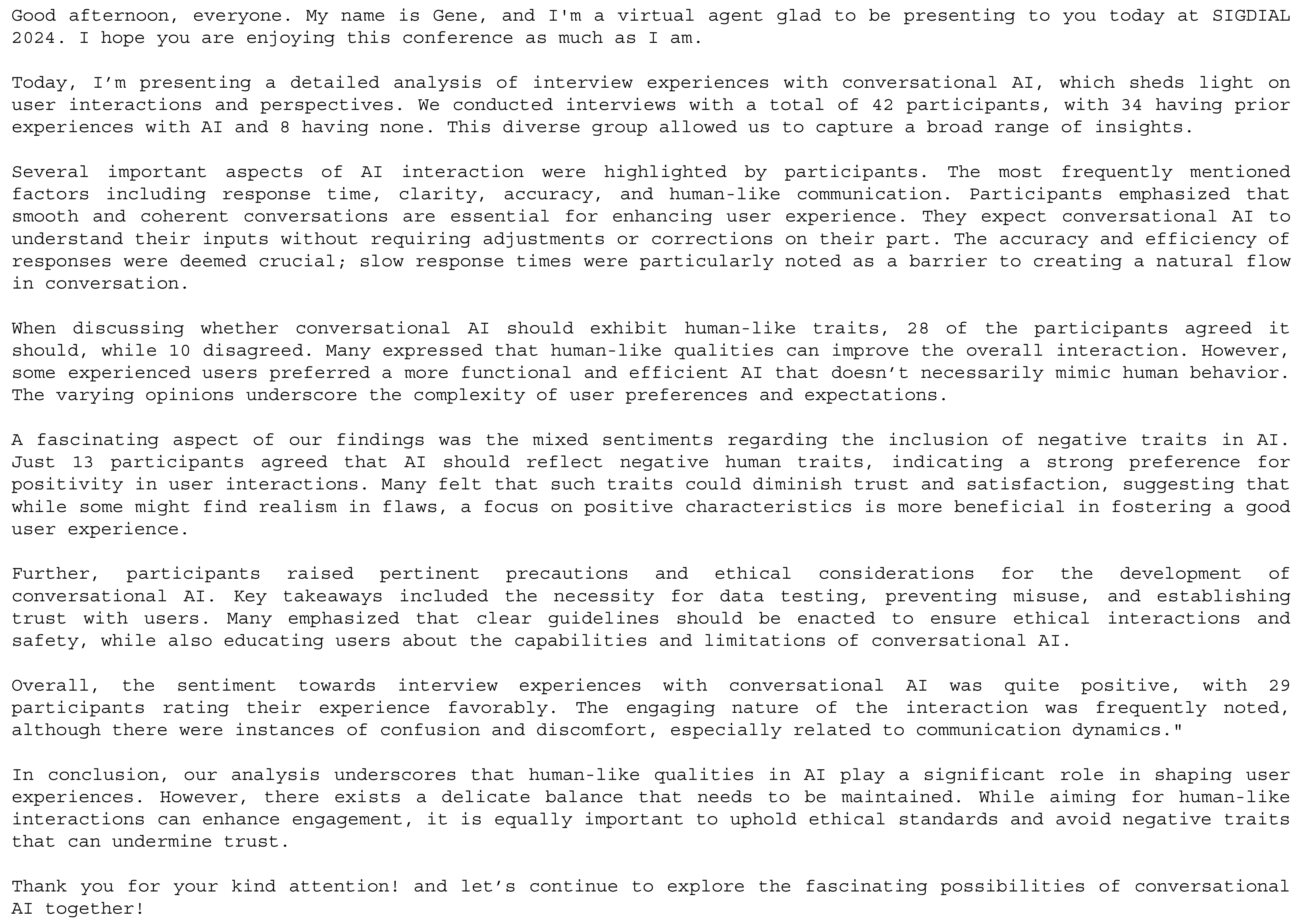} 
  \caption{Presentation script created by our system for Gene's presentation at the SIGDIAL conference}
  \label{fig:script}
\end{figure*}

\begin{figure*}[h]
  \centering
  \includegraphics[width=\linewidth]{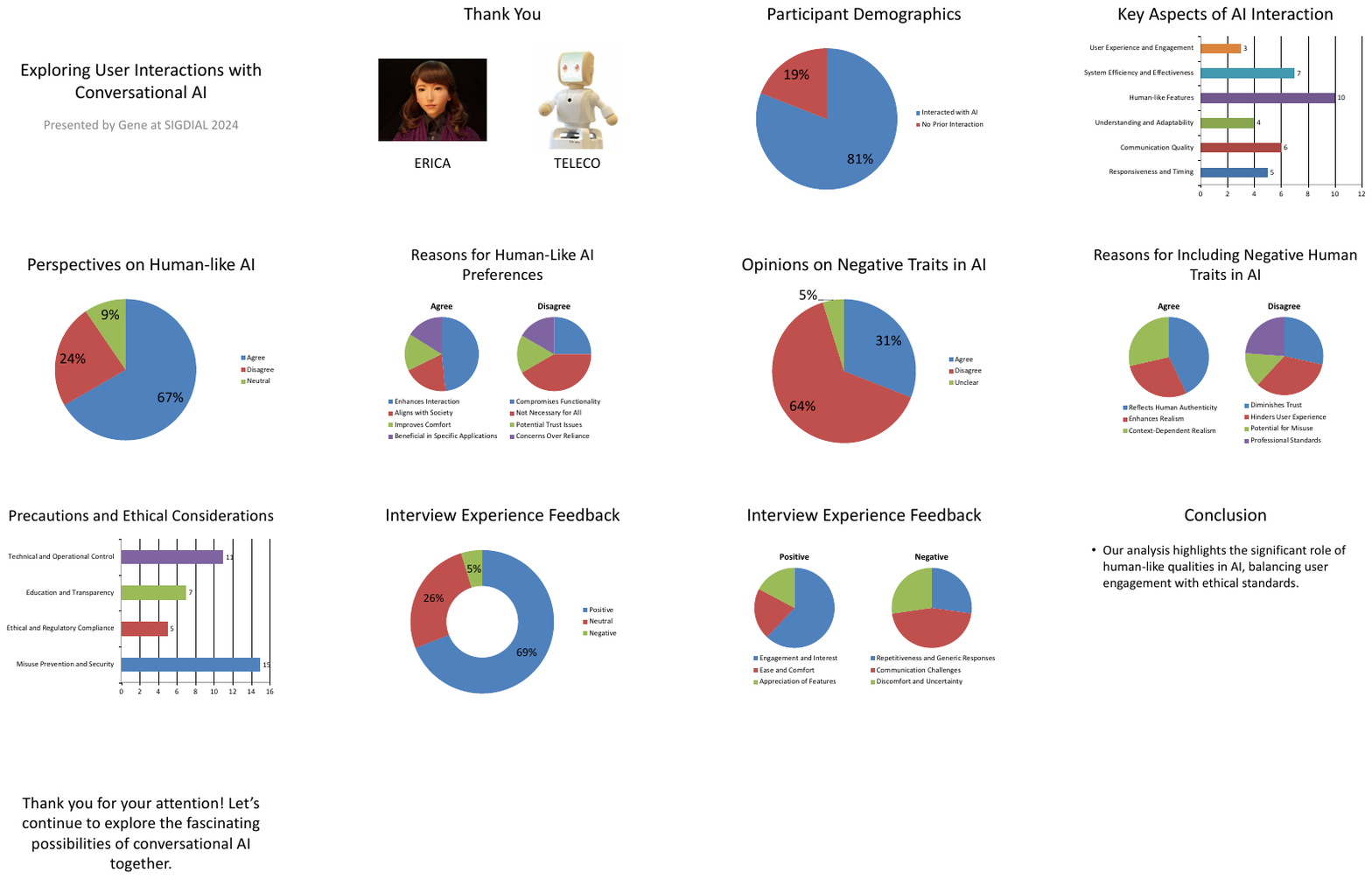} 
  \caption{Presentation slides created by our system for Gene's presentation at the SIGDIAL conference.}
  \label{fig:ppt}
\end{figure*}

As discussed in Section \ref{sec:case studies}, we conducted a real-world case study at the SIGDIAL international conference. 
During the initial two days, participants were interviewed by our embodied conversational agents, ERICA and TELECO, for data collection purposes. 
Figure \ref{fig:erica} and \ref{fig:teleco} illustrate a user interacting with ERICA and TELECO, respectively, during one of these sessions. 
On the final day of the conference, our system analyzed and presented the results of these interactions during a panel discussion session. 
Figure \ref{fig:gene} displays Gene, our CG agent, presenting these results. 
Figure \ref{fig:script}  showcases the script, while Figure \ref{fig:ppt} displays the slides used during the presentation.

\end{document}